\documentclass[10pt,twocolumn,letterpaper]{article}

\usepackage{cvpr}
\usepackage{times}
\usepackage{epsfig}
\usepackage{graphicx}
\usepackage{amsmath}
\usepackage{amssymb}
\usepackage{algorithm}
\usepackage{algorithmic}
\usepackage{dsfont}
\usepackage{array}
\usepackage{wrapfig}
\usepackage[labelformat=simple]{subcaption}

\usepackage[pagebackref=true,breaklinks=true,letterpaper=true,colorlinks,bookmarks=false]{hyperref}

\cvprfinalcopy 

\def\cvprPaperID{618} 

\ifcvprfinal\fi
\begin{document}

\title{Feedback-prop:\,Convolutional Neural Network Inference under Partial Evidence}

\author{Tianlu Wang$^1$,\;\; Kota Yamaguchi$^2$,\;\; Vicente Ordonez$^1$\\
$^1$University of Virginia, $^2$CyberAgent, Inc.\\
{\tt\small yamaguchi\_kota@cyberagent.co.jp}
\\{\tt\small \{tw8cb, vicente\}@virginia.edu}
}

\maketitle

\begin{abstract}
    We propose an inference procedure for deep convolutional neural networks (CNNs) when partial evidence is available. Our method consists of a general feedback-based propagation approach (feedback-prop) that boosts the prediction accuracy for an arbitrary set of \emph{unknown} target labels when the values for a non-overlapping arbitrary set of target labels are \emph{known}. We show that existing models trained in a multi-label or multi-task setting can readily take advantage of feedback-prop without any retraining or fine-tuning. Our feedback-prop inference procedure is general, simple, reliable, and works on different challenging visual recognition tasks. We present two variants of feedback-prop based on layer-wise and residual iterative updates. We experiment using several multi-task models and show that feedback-prop is effective in all of them. Our results unveil a previously unreported but interesting dynamic property of deep CNNs. We also present an associated technical approach that takes advantage of this property for inference under partial evidence in general visual recognition tasks.
\end{abstract}

\vspace{-0.3in}
\section{Introduction}

\begin{figure}[t]
\centering
\vspace{0.3in}
\includegraphics[width=0.46\textwidth]{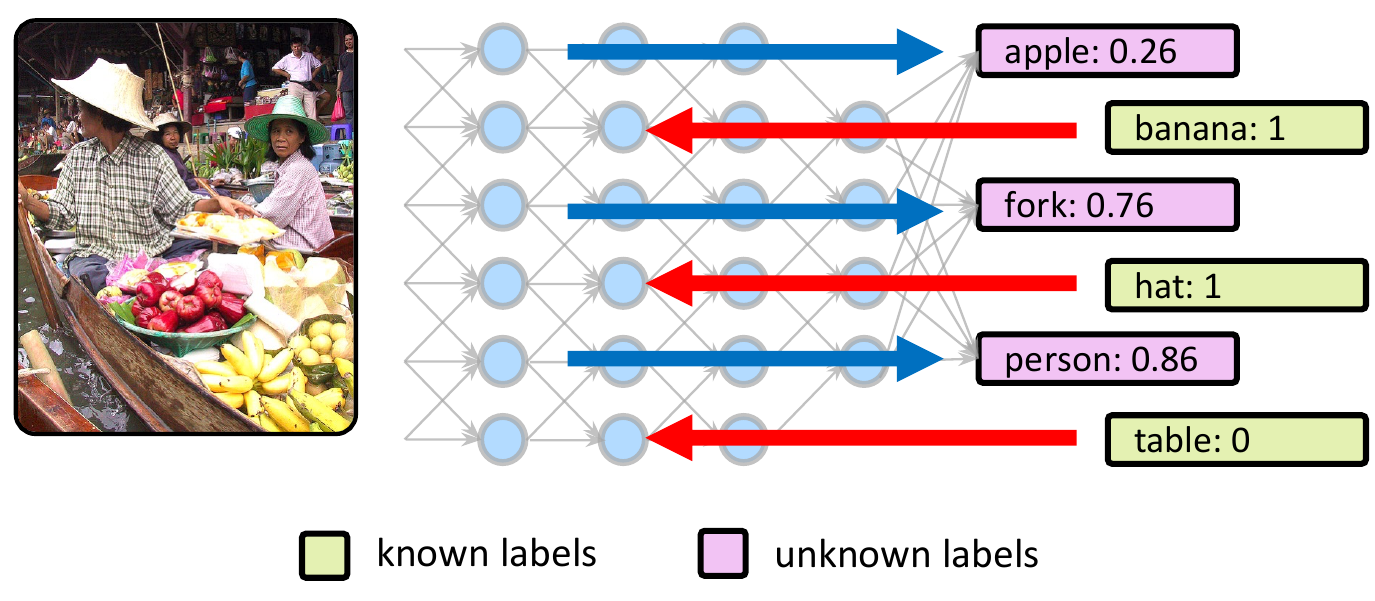}
\caption{Feedback-prop inference leverages an arbitrary set of \emph{known} labels to iteratively predict a set of \emph{unknown} labels for a test input image. This example shows a multi-label classification task. Neural activations are used to transfer information among variables in the target space.}
\vspace{-0.17in}
\label{fig:leadfigure}
\end{figure}

 In this paper we tackle visual recognition problems where partial evidence or partial information about an input image is available at test time. For instance, if we know for certain that an image was taken at the \emph{beach}, this should change our beliefs about the types of objects that could be present, e.g. an \emph{office chair} would be unlikely. This is because something is \emph{known} for certain about the image even before performing any visual recognition. We argue that this setting is realistic in many applications. For instance, images on the web are usually surrounded by text, images on social media have user comments, many images contain geo-location information, images taken with portable devices contain other sensor information. More generally, images in standard computer vision datasets are effectively partially annotated with respect to a single task or modality. Assuming only visual content as inputs, while convenient for benchmarking purposes, does not reflect many end-user applications where extra information is available during inference. We propose here a general framework to address this problem in any task involving deep convolutional neural networks trained with multiple target outputs (i.e.~multi-label classification) or multiple tasks (i.e.~multi-task learning). We provide an example in Figure~\ref{fig:leadfigure}, where a set of labels are \emph{known}: \texttt{banana}, \texttt{hat}, \texttt{table}, while we are trying to predict the other labels: \texttt{apple}, \texttt{fork}, \texttt{person}.

Convolutional neural networks (CNNs) have become the state-of-the-art in most visual recognition tasks. Their extraordinary representation ability has allowed researchers to address problems at an unprecedented scale with remarkable accuracy. While reasoning under partial evidence using probabilistic graphical models would involve marginalization over the variables of interest, CNNs do not model a joint distribution, therefore making such type of reasoning non-trivial. The typical pipeline using CNNs for visual recognition involves training the model using stochastic gradient descent (SGD) and the back-propagation algorithm~\cite{rumelhart} using an annotated image dataset, and then performing forward-propagation during inference given only visual input. In this paper, we challenge this prevailing inference procedure in CNNs where information only flows in one direction, and the model structure is static and fixed after training. We propose instead feedback-based propagation (feedback-prop) where forward and backward-propagation steps use intermediate neural activations to share information among output variables during inference.
We show the effectiveness of our approach on multi-label prediction under incomplete and noisy labels, hierarchical scene categorization, and multi-task learning with object annotations and image descriptions.

Our main hypothesis is that by \emph{correcting} an intermediate set of neural activations using partial labels for a given input sample, we would also be able to make more accurate predictions for the complement set of \emph{unknown} labels. We demonstrate this behavior using our feedback-prop inference for multiple tasks and under multiple CNN models. There is remarkable evidence in previous research aimed at interpreting intermediate representations in CNNs showing that they encode basic patterns of increasing visual complexity (i.e. edges, attributes, object parts, objects) that are shared among target outputs~\cite{simonyan2013deep,zeiler2014visualizing,escorcia2015relationship,vittayakorn2016automatic,netdissect2017}. Since the underlying shared representations of a CNN capture common patterns among target outputs, we find that they can act as pivoting variables to transfer knowledge among variables in the target space. We show that feedback-prop is general, simple to implement, and can be readily applied to a variety of problems where a model is trained to predict multiple labels or multiple tasks. Our code and data are available\footnote{\url{https://github.com/uvavision/feedbackprop}}.
\vspace{0.05in}

Our contributions can be summarized as follows:
\begin{itemize}
\vspace{-0.06in}
    \item A general feedback-based propagation inference procedure (feedback-prop) for CNN inference under partial evidence.
    \vspace{-0.07in}
    \item Two variants of feedback-prop using layer-wise feedback updates, and residual feedback updates, and experiments showing their effectiveness on both multi-label and multi-task settings, including an experiment using in-the-wild web data.
    \vspace{-0.07in}
    \item An extensive analysis of CNN architectures regarding optimal layers in terms of information sharing with respect to target variables using feedback-prop.
\end{itemize}

\section{Related Work}

\paragraph{Use of Context in Computer Vision}
Using contextual cues in visual recognition tasks has long been studied in the psychology literature~\cite{palmer1975effects,navon1977forest,biederman1982scene,chun1998contextual,bar2004visual}, and some of these insights have also been used in computer vision~\cite{rabinovich2007objects,galleguillos2008object,santosh,roozbeh,johnson2015love}. However, unlike our paper, most previous works using context still assume no extra information about images during inference. Instead, contextual information is predicted jointly with target variables, and is often used to impose structure in the target space based on learned priors, label relation ontology, or statistics. In contrast, our work leverages during inference the underlying contextual relations that are already implicitly learned by a CNN.

\vspace{-0.12in}
\paragraph{Conditional Inference in Graphical Models} Our work also has connections with graphical models where messages are iteratively passed through nodes in a learned model that represents a joint distribution~\cite{murphy1999loopy,Salakhutdinov2009DeepBM}.
In our inference method, messages are passed between nodes in a convolutional neural network in forward and backward directions using gradients, intermediate activations, as well as additional residual variables.

\vspace{-0.12in}
\paragraph{Multi-task Learning} Another form of using context is by jointly training on multiple correlated visual recognition tasks or multi-task learning~\cite{Ruan2017Ensemble,Wang2017Attribute,Kokkinos17UberNet}, where knowledge about one task helps another target task. Our inference method is highly complementary and especially useful with these types of models as it can directly be used when extra information is available for at least one of the tasks or modalities. Unlike simple conditional models that would require re-training under a fixed set of conditional input variables, feedback-prop may be used with an arbitrary set of target variables, and does not require re-training.

\vspace{-0.12in}
\paragraph{Optimizing the Input Space}
 In terms of technical approach, feedback-prop has connections to previous works that optimize over inputs. One prominent example is the generation of adversarial examples that are constructed to fool a CNN model~\cite{goodfellow2014explaining}. This style of gradient-based optimization over inputs is also leveraged in the task of image style transfer~\cite{gatys2016image}. Gradients over inputs are also used as the supervisory signal in the generator network of Generative Adversarial Networks (GANs)~\cite{goodfellow2014generative}. Gradient-based optimization has also been used to visualize, identify, or interpret the intermediate representations learned by a deep CNN~\cite{simonyan2013deep,cao2015look,xie2016interactive,zhang2016top,selvaraju2016grad,Binder2016}. However, unlike these methods, we are still interested in the target predictions and not the inputs. 
 We find that CNN layers that lie somewhere in the middle are more beneficial to optimize as pivot variables under our model than the input image.

\vspace{-0.12in}
\paragraph{Deep Inference under Partial Annotations}
In terms of setup, a relevant recent experiment was reported in Hu~et~al~\cite{hu2016learning}. This work introduces a novel deep Structured Inference Neural Network (SINN) model that can be adapted to a setting where true values for a set of labels are \emph{known} at test time. We compare feedback-prop against a re-implementation of SINN for fine-grained scene categorization when a set of coarse scene categories are used as \emph{known} labels, demonstrating superior performance without additional parameters. Tag completion is another relevant problem~\cite{wu2013tag}, but our approach is not specific to multi-label inference and can be easily applied to multiple diverse tasks.

\section{Method}
This section presents our feedback-based inference procedure. We start from the derivation of a basic single-layer \emph{feedback-prop} inference (Sec~\ref{sec:single-layer-feedback}), and introduce our two more general versions: \emph{layer-wise feedback-prop} (LF) (Sec~\ref{sec:layer-wise-feedback}), and our more efficient \emph{residual feedback-prop} (RF) (Sec~\ref{sec:residual-feedback}).

\subsection{Feedback-prop}\label{sec:single-layer-feedback}
Let us consider a feed-forward CNN already trained to predict multiple outputs for either a single task or multiple tasks. Let $\hat{Y} = F(X, \Theta)$ represent this trained CNN, where $X$ is an input image, $\hat{Y}$ is a set of predicted output variables, and $\Theta$ are the model parameters. Now, let us assume that the true values for some output variables are \emph{known} at inference time, and split the variables into \emph{known} and \emph{unknown}: $Y = (Y_{\texttt{k}}, Y_{\texttt{u}})$. The neural network by default makes a joint prediction for both sets of variables: $\hat{Y} = (\hat{Y}_{\texttt{k}}, \hat{Y}_{\texttt{u}}) = (F_{\texttt{k}}(X, \Theta), F_{\texttt{u}}(X, \Theta))$. Given a \emph{known} set of true values $Y_{\texttt{k}}$, we can compute a partial loss only with respect to this set for input sample $X$ as $L(Y_{\texttt{k}}, \hat{Y}_{\texttt{k}})$. The key idea behind feedback-prop is to back-propagate this partially observed loss to the network, and iteratively update the input $X$ in order to re-compute the predictions on the set of \emph{unknown} variables $Y_{\texttt{u}}$. Formally, our basic feedback-based procedure can be described as follows:
\begin{align}
X^* &= \mathrm{argmin}_{X} L(Y_{\texttt{k}}, F_{\texttt{k}}(X, \Theta)), \label{eq:input_feedback} \\
\hat{Y}^*_{\texttt{u}} &= F_{\texttt{u}}(X^*, \Theta),
\end{align}
where we optimize $X$, which acts as our pivoting variable, and forward-propagate to compute refined \emph{unknown} variables $\hat{Y}^*_{\texttt{u}}$. In fact, we need not be restricted to optimize $X$ and can generalize the formulation to optimize arbitrary intermediate representations.
Let us denote the $l$-th layer internal neural activations of the network as $a_l$, and the dissected network at layer $l$ by $Y = F^{(l)}(a_l)$, which can be interpreted as a truncated forward propagation in the original network from layer $l$ until the output. Then, we can define \emph{single-layer feedback-prop} as follows:
\begin{figure}[t]
  \centering
  \includegraphics[width=0.47\textwidth]{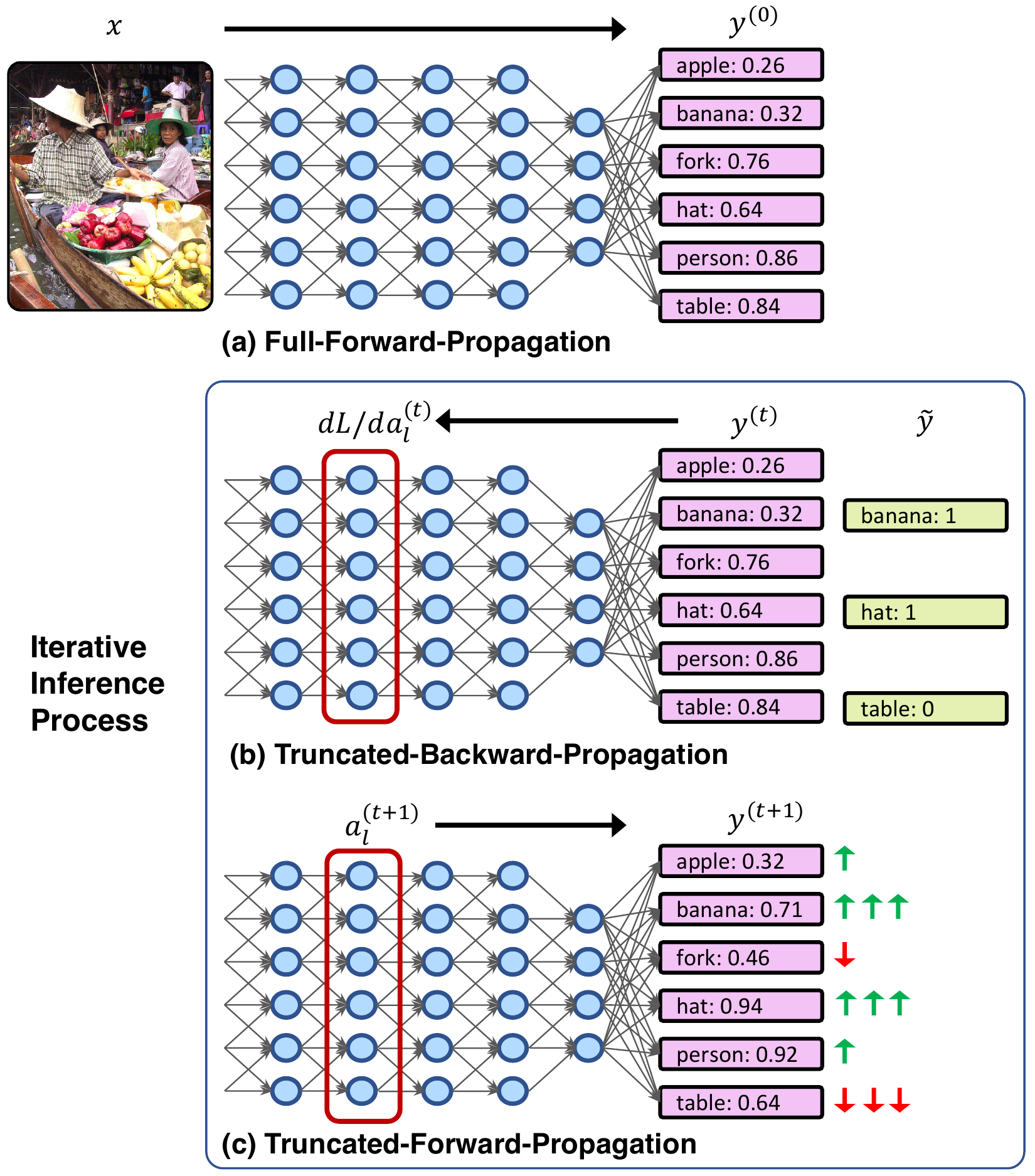}
  \caption{Overview of our feedback-prop iterative inference procedure consisting of three basic steps - (a) full forward propagation to predict initial scores for all labels, (b) truncated backward propagation to update intermediate activations based on the partial evidence (\emph{known} labels), and (c) truncated forward propagation to update the scores for the \emph{unknown} labels.}
  \label{fig:modelfigure}
\end{figure}
\begin{align}
  a_l^* &= \mathrm{argmin}_{a_l} L(Y_{\texttt{k}}, F_{\texttt{k}}^{(l)}(a_l, \Theta)), \label{eq:activation_feedback} \\
  \hat{Y}_{\texttt{u}} &= F_{\texttt{u}}^{(l)}(a_l^*, \Theta).
\end{align}
In this formulation, we optimize intermediate representations at an arbitrary layer in the original model shared by $F_{\texttt{k}}$ and $F_{\texttt{u}}$. These intermediate neural activations act as pivoting variables. Note that equation \ref{eq:input_feedback} is a special case of single-layer feedback-prop when $a_{0} \equiv X$. 

In our description of feedback-prop we define the output space $Y$ as a set of variables. Each output variable can be arbitrarily complex, diverse and seemingly unrelated, as is often the case in multi-task models. In the simpler scenario of multi-label prediction, each variable corresponds to a label. We illustrate in Figure~\ref{fig:modelfigure} an overview of our feedback-prop approach for a multi-label prediction model.

\subsection{Layer-wise Feedback-prop (LF)}\label{sec:layer-wise-feedback}
In this section we propose a more general version of feedback-prop that leverages multiple intermediate representations in a CNN across several layers: Layer-wise feedback-prop. This procedure minimizes a loss function $ L(Y_{\texttt{k}}, F_{\texttt{k}}(A, \Theta))$ by optimizing a set of topologically sorted intermediate activation  $A \equiv \{ a_{i}, a_{i+1}, \cdots, a_{N} \}$ starting at layer $i$. However, in feed-forward models, $a_l$ is needed to compute $a_{l+1}$. This requires optimizing these multiple intermediate representations using layer-by-layer sequential updates. We describe \emph{layer-wise feedback-prop} in detail in Algorithm \ref{alg:layer-wise-feedback}. {Forward} represents a truncated forward propagation from the given input at a certain layer until the output layer, and {Backward} represents a truncated back-propagation of gradients from the output layer to the intermediate pivoting activations. Given an input image $X$, known values for variables $Y_{\texttt{k}}$, and a topologically sorted list of layers $\mathcal{L}$, the algorithm optimizes internal representations $a_l$ in topological order. More generally, these layers do not need to be consecutive. The updates are performed in this fashion so that the algorithm \emph{freezes} activation variable $a_l$ layer-by-layer from the input side, so that after each freeze, the next variable can be initialized to apply feedback updates. In Algorithm \ref{alg:layer-wise-feedback}, $\lambda$ is an update rate and iterative SGD steps are repeated $T$ times. The update operation (line 7) may be replaced by other types of SGD update rules such as SGD with momentum, AdaGrad, or Adam. Note that the backward, and forward propagation steps only go back as far as $a_l$, and do not require a full computation through the entire network. The \emph{single-layer feedback-prop} inference in Sec~\ref{sec:single-layer-feedback} is a special case of \emph{layer-wise feedback-prop} when $|\mathcal{L}| = 1$. The choice of layers will affect the quality of feedback-prop predictions for \emph{unknown} targets.

\renewcommand{\algorithmicrequire}{\textbf{Input:}}
\renewcommand{\algorithmicensure}{\textbf{Output:}}
\begin{algorithm}[t!]
  \caption{Layer-wise Feedback-prop Inference}
  \begin{algorithmic}[1]
    \REQUIRE Input image $X$, \emph{known} labels ${Y_{\texttt{k}}}$, and a list of layers $\mathcal{L} \equiv \{{i}, {i+1}, \cdots , {N}\}$
    \ENSURE Prediction $\hat{Y}_{\texttt{u}}$
    \STATE $a_{0}^{(T)} := X$
	\FOR{$l \in \mathcal{L}$}
      \STATE $\hat{Y}_{\texttt{k}}^{(0)}, a_{l}^{(0)} := \text{Forward}(a_{l-1}^{(T)})$
	  \FOR{ $t = 0$ \TO $T$}
	    \STATE Compute the partial loss $L({Y_{\texttt{k}}}, \hat{Y}_{\texttt{k}}^{(t)})$
        \STATE $\frac{\partial L}{\partial a_l^{(t)}} := \text{Backward}(L)$ 
        \STATE $a_l^{(t+1)} := a_l^{(t)} - \lambda \frac{\partial L}{\partial a_l^{(t)}}$
        \STATE $\hat{Y}_{\texttt{k}}^{(t+1)} := \text{Forward}(a_{l}^{(t+1)})$
      \ENDFOR
    \ENDFOR
    \STATE $\hat{Y}_{\texttt{u}} = \text{Forward}(a_N^{(T)})$
  \end{algorithmic}
  \label{alg:layer-wise-feedback}
\end{algorithm}

\subsection{Residual Feedback-prop (RF)}\label{sec:residual-feedback}

The proposed \emph{layer-wise feedback-prop} (LF) inference can use an arbitrary set of intermediate layer activations, but is inefficient due to the double-loop in Algorithm \ref{alg:layer-wise-feedback}, where layers have to be updated individually in each pass. Here, we refine our formulation even further by updating multiple layer activations in a single pass through the incorporation of auxiliary residual variables. We name this version of our inference procedure \emph{residual feedback-prop} (RF) inference.

\begin{figure}[b]
  \centering
  \includegraphics[width=0.44\textwidth]{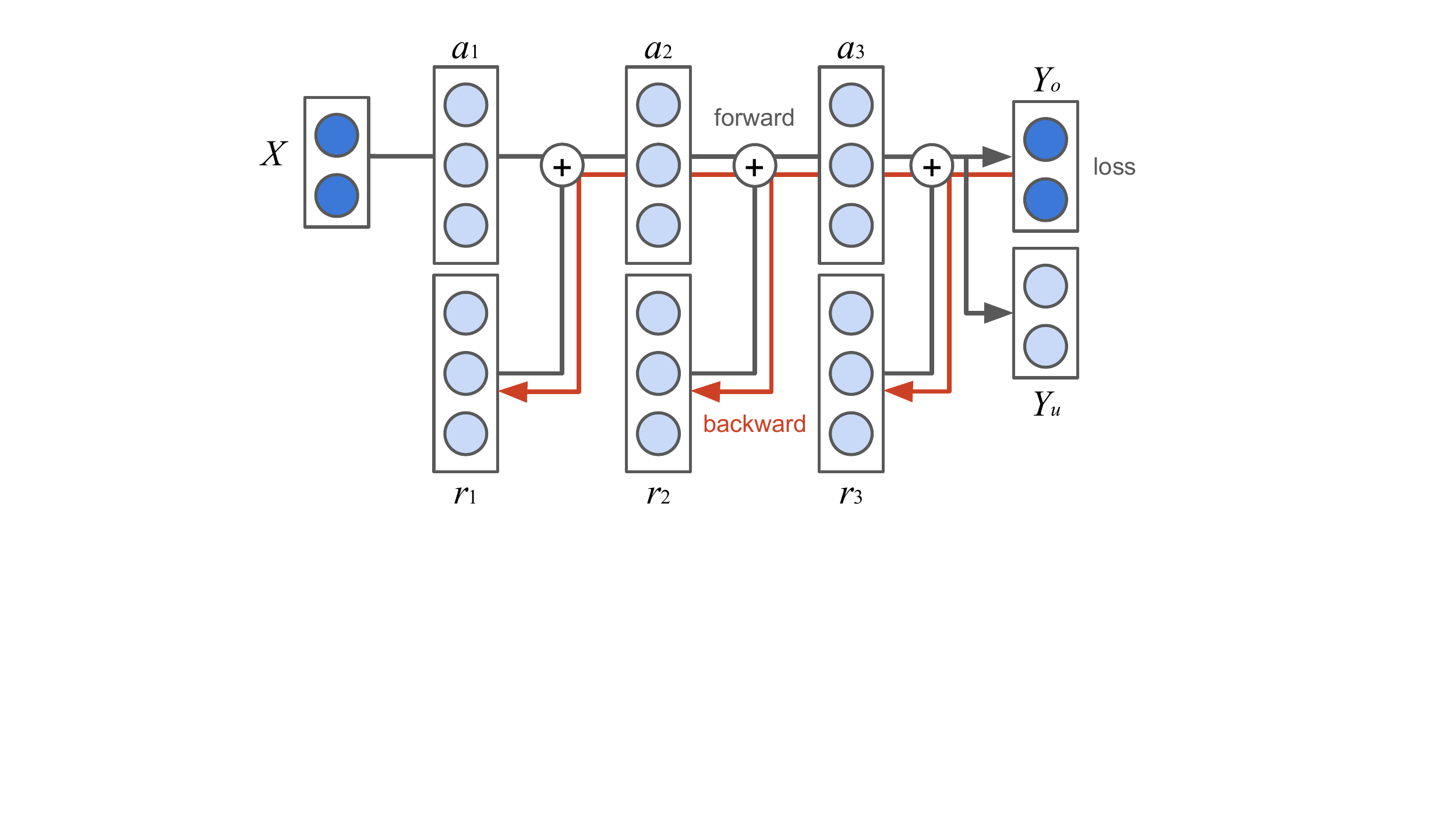}
  \caption{In our RF approach, residual variables $r_l$ are updated instead of intermediate activations $a_l$ in order to update all layers in a single pass.}
  \label{fig:residual}
\end{figure}

The core idea in RF is to inject an additive variable (feedback residual) to intermediate representation variables, and optimize over residuals instead of directly updating intermediate representations. Notice that incorporation of these residual variables takes place only during inference, and does not involve any modifications in learning, or whether the underlying model itself uses residuals. We add a feedback residual variable $r_l$ to the unit activation $a_l$ in the forward propagation at layer $l$ as follows:
\begin{align}
  a_{l} = f_{l}(a_{l-1}, \theta_l) + r_l,
\end{align}
where $f_l$ is the layer transformation function at $l$ (e.g. convolutional filtering)  with model parameters $\theta_l$. When $r_l = 0$, this is a regular forward-propagation. Instead of directly updating $a_l$ by feedback-prop as in LF, we only update residual variables $r_l$. Figure~\ref{fig:residual} shows how residual variables are incorporated in a model during inference. 

\begin{algorithm}[t]
  \caption{Residual Feedback-prop Inference}
  \begin{algorithmic}[1]
    \REQUIRE Input image $X$, \emph{known} labels ${Y_{\texttt{k}}}$, and a list of layers $\mathcal{L} \equiv \{ {i}, {i+1}, \cdots ,N\}$
    \ENSURE Prediction $\hat{Y}_{\texttt{u}}$
    \STATE $\mathbf{r}^{(0)} \equiv \{ r_{l}^{(0)} | l \in \mathcal{L} \} := \mathbf{0}$
    \STATE $a_{0} := X$
    \FOR{ $t = 0$ \TO $T$}
      \FOR{$l \in \mathcal{L}$}
        \STATE $a_l^{(t)}$ := $\text{Forward}(a_{l-1}^{(t)}) + r_{l}^{(t)}$
      \ENDFOR
      \STATE $\hat{Y}_{\texttt{k}}^{(t)} := \text{Forward}(a_{N}^{(t)})$
      \STATE Compute the partial loss $L({Y_{\texttt{k}}}, \hat{Y}_{\texttt{k}}^{(t)})$
      \STATE $\frac{\partial L}{\partial \mathbf{r}^{(t)}} := \text{Backward}(L)$ 
      \STATE $\mathbf{r}^{(t+1)} := \mathbf{r}^{(t)} - \lambda \frac{\partial L}{\partial \mathbf{r}^{(t)}}$
    \ENDFOR
    \STATE $\hat{Y}_{\texttt{u}} = \text{Forward}(a_N^{(T)})$
  \end{algorithmic}
  \label{alg:residual-feedback}
\end{algorithm}

Algorithm \ref{alg:residual-feedback} describes in detail how residual feedback-prop operates. The procedure starts by setting residuals to zero (line 1). The inner-loop is a truncated feed-forward propagation starting in activation $a_l$ but using additive residuals. Notice that this computation does not incur significant computational overhead compared to regular forward propagation. Updates do not require a double-loop (lines 9-10), therefore avoiding repetitive gradient computations as in LF. We show in our experiments that residual-based feedback-prop performs comparably to layer-wise feedback-prop in multi-label and multi-task models, and is more efficient when updating multiple layers (Sec~\ref{sec:residual-efficient}).

\begin{figure*}[th!]
    \centering
    \vspace{-10pt}
    \begin{subfigure}[b]{0.42\textwidth}
        \includegraphics[width=\linewidth]{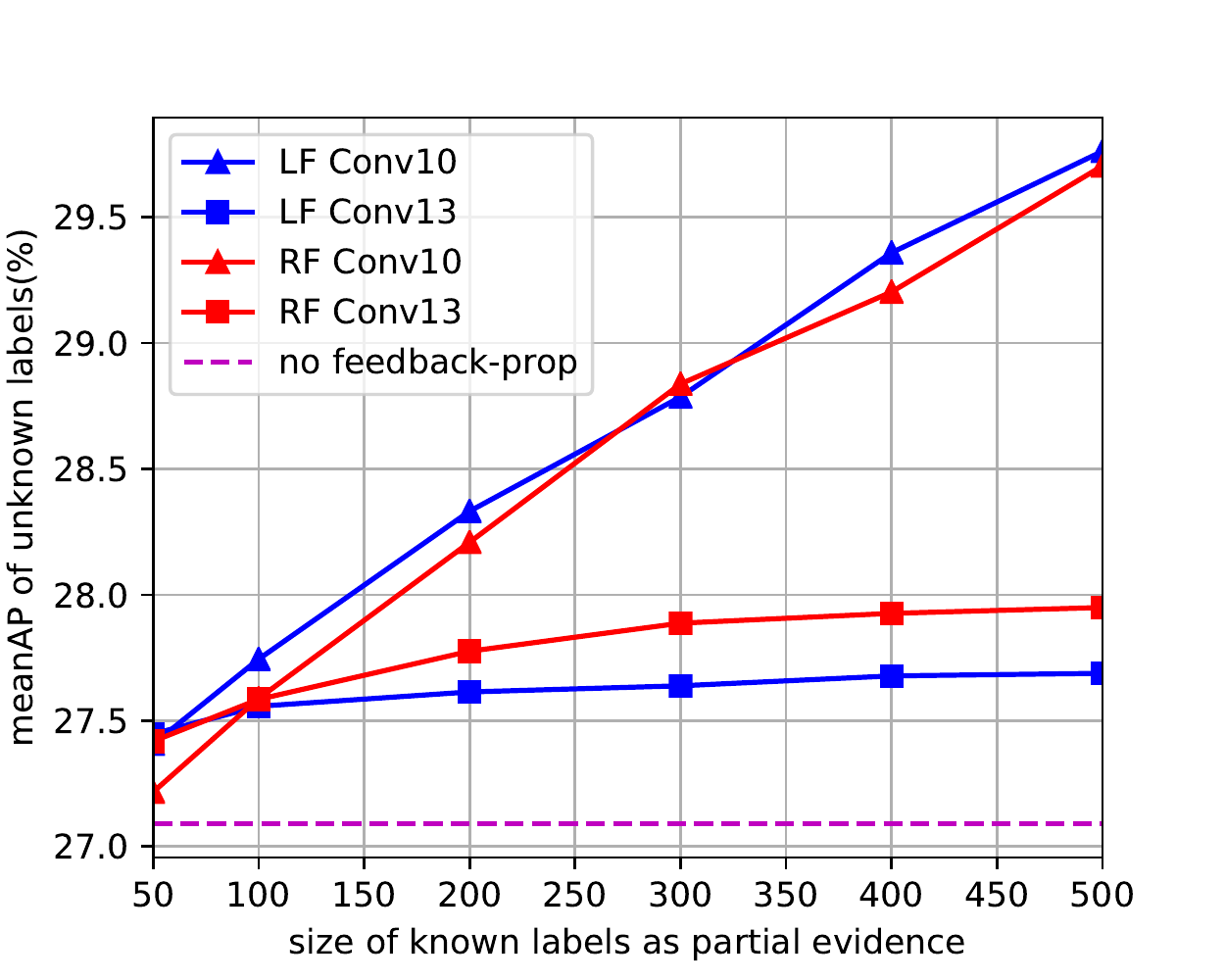}
        \caption{Feedback-prop on VGG16}
        \label{fig:vgg16bn}
    \end{subfigure}
    \quad\quad
    \begin{subfigure}[b]{0.42\textwidth}
        \includegraphics[width=\linewidth]{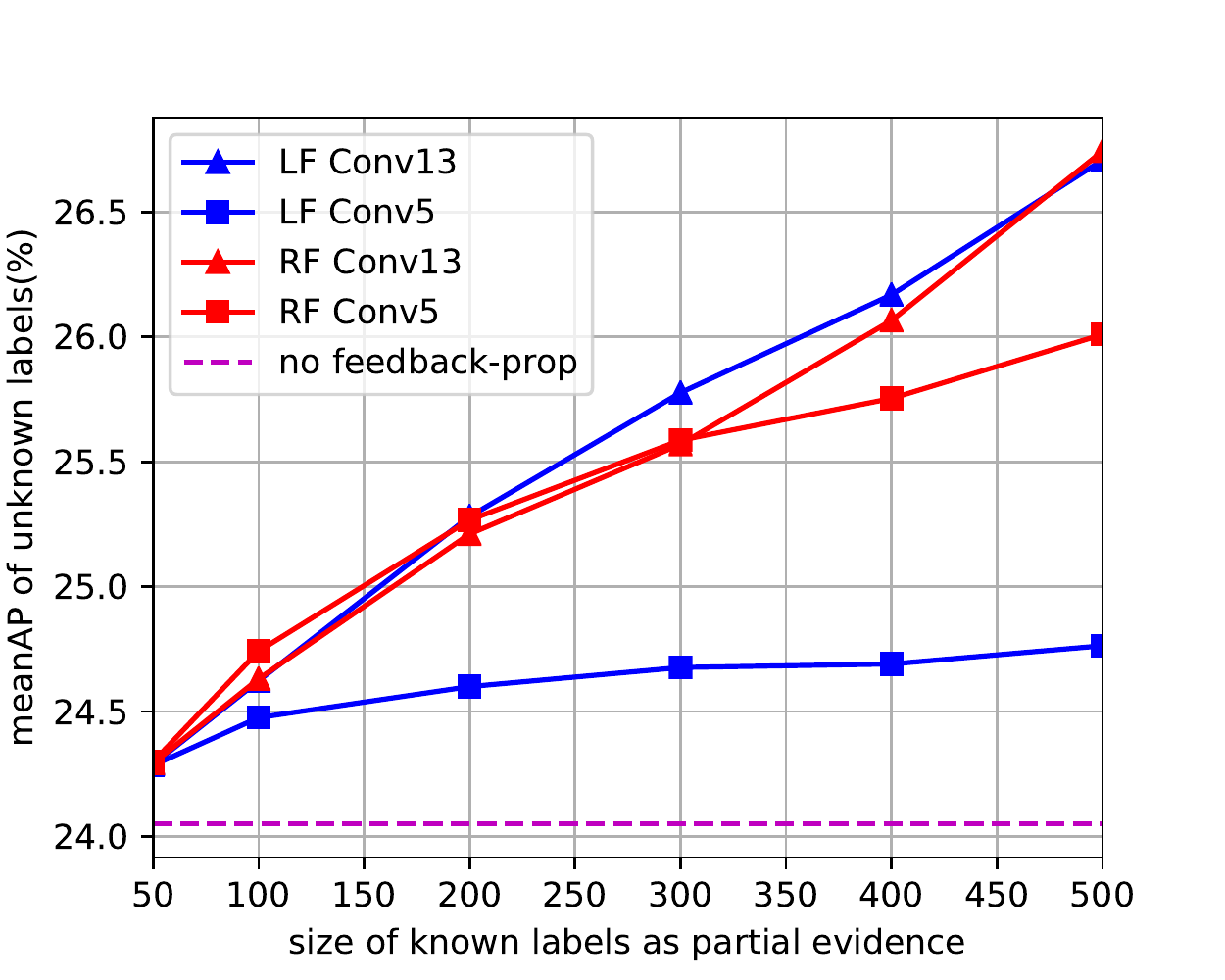}
        \caption{Feedback-prop on ResNet18}
        \label{fig:resnet18}
    \end{subfigure}
    
    \caption{Performance (mAP) of LF and RF using different intermediate activations (Conv5, 10, 13) against the amount of \textit{known} labels in the COCO multi-label image annotation task: the more the labels, the higher the performance.}
    \label{fig:resnet&vgg}
    \vspace{-0.12in}
\end{figure*}

\section{Experiments} \label{sec:experiments}
We evaluate our approach on four tasks 1) Multi-label image annotation with incomplete labels, where incomplete labels are simulated at test time by artificially splitting the total vocabulary of labels into \emph{known} and \emph{unknown} (Sec~\ref{sec:multi-label-experiment}), 2) Hierarchical scene categorization, where true values for coarse scene categories are known and the aim is to predict fine-grained scene categories (Sec~\ref{sec:scene-categorization}), 3) Automatic annotation of news images in-the-wild, where surrounding news text is \emph{known}, and a set of visual words from image captions are the \emph{unknown} targets (Sec~\ref{sec:news-experiment}), and 4) A multi-task joint prediction of image captions and object categories, where the goal during inference is to predict image captions as the \emph{unknown} target (Sec~\ref{sec:multi-task}).

\subsection{Multi-label Image Annotation} \label{sec:multi-label-experiment}
This experiment uses the COCO dataset~\cite{mscoco}, containing around $120\text{k}$ images, each with $5$ human-annotated captions. We use the standard split in the dataset that has $82,783$ images in the training set and subdivide the standard validation set into $20,000$ images for validation and $20,504$ for testing. Our task is to predict visual concepts for any given image similar to the visual concept classifier used by Fang~et~al~\cite{fang2015captions}, which we use as our baseline. We build a vocabulary of concepts using the most frequent $1000$ words in captions from the training set after tokenization, lemmatization, and stop-word removal. We first train a multi-label prediction model by modifying a standard CNN to generate a 1000-dimensional output, and learn logistic regressors using the following loss function:
\begin{equation}
\begin{aligned}
\label{eq:logistic}
L = -\sum_{i=1}^{d}\frac{1}{N}&\sum_{j=1}^{N}\lambda_j[ 
   y_{ij}\log\sigma(f_j(I_i,\Theta))\; + \\
 &  (1 - y_{ij})\log(1 - \sigma(f_j(I_i,\Theta)))],
\end{aligned}
\end{equation}
\noindent where $\sigma(x)=1 / (1+exp(-x))$ is the sigmoid function, $f_j(I_i, \Theta)$ is the unnormalized output score for category $j$ given image $I_i$, and $\Theta$ are the model parameters of the underlying CNN. Intuitively, each term in this loss function encourages activation $f_j$ to increase if label $y_{ij}=1$ or decrease otherwise. Weight parameters $\lambda_j$ count the contribution of each class $j$ differently. These parameters are designed to handle the extreme class imbalance in multi-label image annotation - larger values of lambda are assigned to classes that occur less frequently. Particularly, we set $\lambda_j = {\sum_{i=1}^{|D|}(1 - y_{ij})}\,/\,{\sum_{i=1}^{|D|}y_{ij}}.$ We load weights from models pretrained on ImageNet to train our models.

For feedback-prop evaluation, we put aside a fixed set of $500$ targets as \emph{unknown}. We measure the mean average precision, mAP, (area under the precision-recall curve) averaged on the \emph{unknown} label set as we experiment with different amounts of \emph{known} labels, from $50$ to the total complement set of $500$ labels.  Figure~\ref{fig:resnet&vgg} reports the results for both LF and RF, using several intermediate representations from VGG-16~\cite{vgg} and Resnet-18~\cite{resnet}. We determine the update rate parameter and number of iterations using the validation split, and report results on the test split. When the amount of \emph{known} labels is less than $500$, we run $5$ rounds with randomly sampled labels and report average performance.

\noindent\textbf{Observations:} Remarkably, for both LF and RF, accuracy increases with the amount of partial evidence without any apparent diminishing returns.
Different layers achieve different levels of accuracy, indicating that information shared with the target label space changes across internal convolutional layers in both Resnet-18 and VGG-16. Figure \ref{fig:vgg16bn} shows that VGG-16 achieves a mAP on the set of \textit{unknown} labels of $27.09$ when using only the image as input, and the mAP is improved to 27.41 on average when only using a random sample of $50$ \emph{known} labels when using the outputs of Conv13 as pivoting variables under LF. Note that these $50$ \emph{known} labels are potentially unrelated to the $500$ labels the model is trying to predict, and most of them only provide weak negative evidence (e.g. $y_{ij}$ = 0). When using the full complement set of $500$ labels, the predictions achieve $29.76$ mAP, which represents a 9.8\% relative improvement. Figure \ref{fig:resnet18} shows that Resnet-18 achieves a mAP of $24.05$ using no additional evidence. RF under Conv13 outputs as pivoting variables can reach $26.74$ mAP given the non-overlapping set of $500$ \textit{known} labels as partial evidence, a relative improvement of 11.2\%.

\subsection{Hierarchical Scene Categorization} \label{sec:scene-categorization}

We apply feedback-prop on scene categorization on the SUN dataset~\cite{xiao2010sun}. This dataset has images annotated with 397 fine-grained scene categories, 16 general scene categories, and 3 coarse categories. We follow the same experimental setting of train, validation and test split ratio reported in \cite{Agrawal2014AnalyzingTP} with $50$, $10$ and $40$ images from every scene category. Our task is to infer fine-grained categories given true values for coarse categories as it was performed in Hu et~al~\cite{hu2016learning}. For evaluation, we compute multi-class accuracy (\textit{MC Acc}) and intersection-over-union accuracy (\textit{IoU Acc}) as well as mean average precision (\textit{$mAP$}) averaged over all categories. 

\begin{table}[h]
\centering
\setlength\tabcolsep{3.6pt}
    \begin{tabular}{ p{2.15cm}|c|c|c  }
     
     & \textit{MC Acc} & \textit{$mAP$} & \textit{IoU Acc}\\
     \hline
    Baseline~\cite{hu2016learning} & 52.83$\pm$0.24 & 56.17$\pm$0.21 & 35.90$\pm$0.22 \\
    \hline
    Bsln\,+\,PL~\cite{hu2016learning}& 53.15$\pm$0.27 & 56.49$\pm$0.24 & 36.20$\pm$0.26 \\
    SINN\,+\,PL~\cite{hu2016learning}& 54.30$\pm$0.35 & 58.45$\pm$0.31 & 37.28$\pm$0.34\\
    Ours (LF) & 54.93$\pm$0.42 & 58.52$\pm$0.34 & 37.86$\pm$0.39\\
    Ours (RF) & \bf{55.01$\pm$0.35} & \bf{58.70$\pm$0.26} & \bf{37.95$\pm$0.33} \\
     \hline
    \end{tabular}
    \caption{
    Feedback-prop on hierarchical scene categorization in SUN397. Our methods (LF / RF) outperform baseline methods on all metrics when partial labels are available.}
     \label{tab:sun397}
     \vspace{-0.12in}
\end{table}

\noindent\textbf{Observations:} Table~\ref{tab:sun397} reports results averaged over 5 runs. We use a CNN + Softmax classifier as our first Baseline, and as a second baseline a CNN + Softmax classifier that uses true values for coarse categories in the form of a binary indicator vector as additional input to the classifier (Baseline\,+\,PL). Similar baselines were used in Hu~et~al~\cite{hu2016learning}. Additionally, we re-implement the Structured Inference Neural Network (SINN) of Hu~et~al~\cite{hu2016learning} which outputs three levels of predictions for fine-grained, general, and coarse scene categories and connects them using a series of linear layers modeling positive and negative relations in the target space and in both top-down and bottom-up directions. Instead of using WordNet to estimate label relations, we threshold pearson correlation coefficients between target variables in the training split. 
Both LF and RF successfully outperform the baselines and the previously proposed model in all metrics. Notice that our proposed method does not require a significant amount of additional parameters. In these experiment RF and LF use as pivoting variables the outputs of Conv-\{2, 3, 4, 5\}. For this experiment, all models rely on Alexnet~\cite{alexnet} pretrained in the Places365 dataset~\cite{zhou2017places}. 

\subsection{Visual Concept Prediction on News Images}
\label{sec:news-experiment}
In this experiment, we train a multi-task model that jointly predicts a set of visual concepts from news image captions and a separate set of concepts from surrounding text. We first collected a dataset of news images with associated captions and text from the BBC news website. Our splits have $153,364$ images for training, $10,213$ images for validation, and $10,307$ images for testing. Both tasks are trained under the same multi-label loss and setup from Sec~\ref{sec:multi-label-experiment}. The vocabulary for visual concepts from image captions consists of the $500$ most frequent nouns, and the vocabulary for visual concepts from surrounding news texts consists of the top $1,000$ most frequent nouns. We use Resnet-50~\cite{resnet} trained under the sum of the losses for each task. At inference time, we predict the visual concepts defined by words in captions (\emph{unknown} labels), given the input image and the surrounding news text (\emph{known} labels). We evaluate LF using layer Conv40 and RF under Conv22 as pivoting variables respectively, which we generally find to perform best in previous experiments.
Table~\ref{tab:bbcnews} shows the mAP across the set of  \emph{unknown} labels in the test split with varying amounts of additional partial evidence (surrounding news text).

\newcolumntype{x}[1]{>{\centering\arraybackslash\hspace{0pt}}p{#1}}

\begin{table}
\centering
    \begin{tabular}{ p{2cm}| x{1.8cm} | x{1.8cm} }
     
      & LF-conv-40 & RF-conv-22 \\
     \hline
     no-text & 19.92 & 19.92 \\
     \hline
     25\% text & 21.33 & 21.27  \\
     50\% text & 22.16 & 22.23 \\
     75\% text & 22.42 & 22.51  \\
     100\% text & \textbf{22.57} & \textbf{22.57}  \\
     \hline
    \end{tabular}
    \caption{
    mAP of visual concept predictions on news images without vs with surrounding news text.
    }
     \label{tab:bbcnews}
     \vspace{-0.23in}
\end{table}

\begin{figure*}[t!]
\centering
\includegraphics[width=0.98\textwidth]{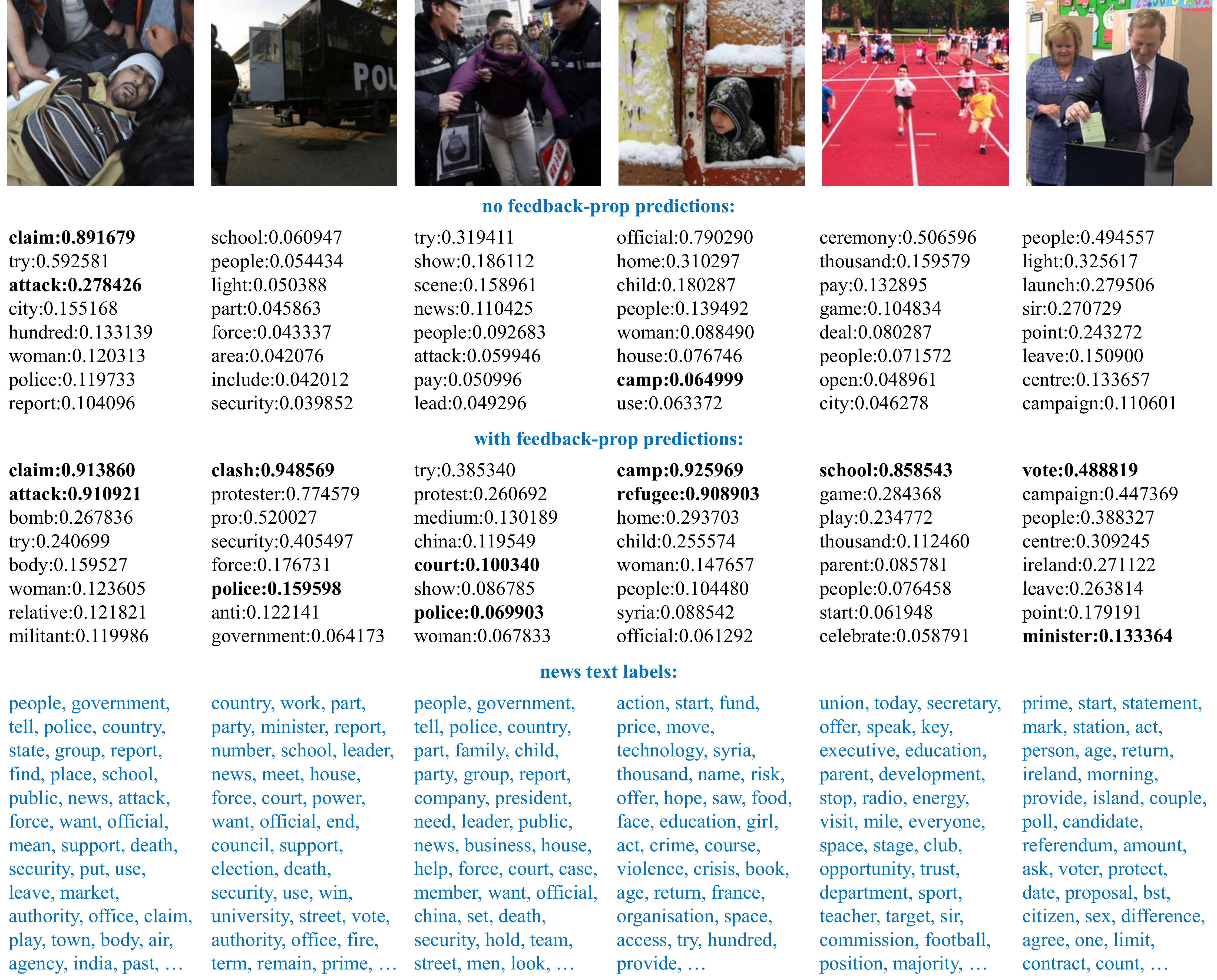}
\caption{Qualitative examples for visual concept prediction for News Images. Second row shows results of a multi-label prediction model (no feedback-prop), the next row shows results obtained using LF where words from surrounding news text (shown in blue) are used as partial evidence. Predictions also among the true labels are highlighted in bold. While news text contains many words that seem marginally relevant, feedback-prop still leverages them effectively to improve predictions. Surrounding news text provides high-level feedback to make predictions that would otherwise be hard.}
\label{fig:qualitative}
\end{figure*}

\noindent\textbf{Observations:}  The mAP for predicting the set of \emph{unknown} labels improves from $19.921\%$ (only using input images) to $21.329\%$ even when only using the first $25\%$ of the surrounding news text as additional evidence. Using a larger portion of surrounding news text consistently increases the accuracy. When using all the available surrounding text for each news image the mAP improves on average from $19.92\%$ to $22.57\%$, a relative improvement of 13.3\%. This is remarkable since --unlike our previous experiment-- the surrounding text might also contain many confounding signals and noisy labels. We show qualitative examples of LF using all surrounding text as partial evidence in Figure~\ref{fig:qualitative}.

\begin{table*}
\parbox{.32\linewidth}{
\centering
    \begin{tabular}{ p{1.6cm}|x{1.3cm} | x{1.3cm}  }
      & LF & RF\\
     \hline
     no-fp & 26.98 & 26.98\\
     \hline
     fp-input & 29.14 &29.53 \\
     fp-conv-1 &29.72 &29.56\\
     fp-conv-4 &29.65 &29.66 \\
     fp-conv-7  &29.77 &\textbf{29.79} \\
     fp-conv-10 &\textbf{29.82} &29.74 \\
     fp-conv-13 &27.59 &27.87\\
     \hline
    \end{tabular}
    \caption{VGG-16 layer-wise analysis.}
     \label{tab:layer-analysis-3}
}
\hfill
\parbox{.32\linewidth}{
\centering
\centering
    \begin{tabular}{ p{1.6cm}|x{1.3cm} | x{1.3cm} }
     
      & LF & RF \\
     \hline
     no-fp & 24.08 & 24.08 \\
     \hline
     fp-input & 24.74 & 27.06  \\
     fp-conv-1 & 24.16 & 25.91 \\
     fp-conv-5 & 24.57 & 25.76  \\
     fp-conv-9 & 25.94 & 26.71  \\
     fp-conv-13 & \textbf{26.80} & \textbf{27.26}  \\
     fp-conv-17 & 24.19 & 24.22 \\
     \hline
    \end{tabular}
    \caption{
    Resnet-18 layer-wise analysis.}
     \label{tab:layer-analysis-1}
}
\hfill
\parbox{.32\linewidth}{
\centering
    \begin{tabular}{ p{1.6cm}|x{1.3cm} | x{1.3cm}  }
      & LF & RF\\
     \hline
     no-fp & 26.94 & 26.94\\
     \hline
     fp-input &28.35 &29.28 \\
     fp-conv-1 &27.60 &29.49\\
     fp-conv-10 &29.54 &29.80 \\
     fp-conv-22  &29.61 &\textbf{29.89} \\
     fp-conv-40 &\textbf{29.71} &29.67 \\
     fp-conv-49 &27.14 &27.14\\
     \hline
    \end{tabular}
    \caption{
    Resnet-50 layer-wise analysis.}
     \label{tab:layer-analysis-2}
}
\vspace{-0.1in}
\end{table*}

\subsection{Joint Captioning and Object Categorization}
\label{sec:multi-task}
We train a multi-task CNN model on the COCO dataset~\cite{mscoco} to jointly perform caption generation and multi-label object categorization. We use Resnet-50 with two additional output layers after the last convolutional layer: a multi-label prediction layer with $80$-categorical outputs corresponding to object annotations, and an LSTM decoder for caption generation as proposed by Vinyals~et~al~\cite{showandtell}. We shuffle images in the standard COCO train and validation splits and use $5000$ images for validation and test, and the remaining samples for training. We perform the same pre-processing on images and captions as in \cite{Karpathy_2015_CVPR}. We report BLEU\cite{bleu}, METEOR\cite{meteor} and CIDEr\cite{cider} scores for captioning and mean average precision(mAP) for object categorization. This model achieves a $0.939$ CIDEr score and $71.3\%$ \emph{mAP}. In order to evaluate feedback-prop, we use object annotations as \emph{known} and analyze the effects on the quality of the predicted captions -- our \emph{unknown} target. 
Table \ref{tab:multi-task} presents results under this regime on the test split. 

\begin{table}[h]
\centering
    \begin{tabular}{ p{2.2cm}|c|c|c  }
     
     &BLEU-4 & ROUGE & CIDEr\\
     \hline 
     no-fp~\cite{showandtell} & 28.65 & 0.5267 & 0.9466\\
     \hline
    LF-input & 29.20 & 0.5290 & 0.9647 \\
    LF-conv-10 & 29.78 & 0.5333 & 0.9859 \\
    LF-conv-22 & 29.71 & 0.5327 & 0.9834 \\
    LF-conv-40 & 29.66 & 0.5332 & 0.9854 \\
    LF-conv-10, 40 & 29.73 & 0.5329 & 0.9872  \\
    RF-conv-10, 40 & 29.63 & 0.5337 & 0.9922\\
     \hline
    \end{tabular}
    \caption{
    Feedback-prop in multi-task learning: caption
    generation results benefit from object annotations as partial evidence using feedback-prop.}
     \label{tab:multi-task}
\end{table}

\noindent\textbf{Observations:} Feedback propagation between target outputs and intermediate representations (including inputs) helps generate better image captions. We observe that using LF with any layer as pivot, improves the predictions under all standard metrics. Furthermore, we observe that jointly using the outputs of layers Conv10 and Conv40 as pivots can outperform updating the outputs of any single layer. RF on Conv10 and Conv40 reaches the highest CIDEr score, improving from $0.946$ to $0.992$. 

\begin{figure}[b]
\centering
\vspace{-0.22in}
\includegraphics[width=0.36\textwidth]{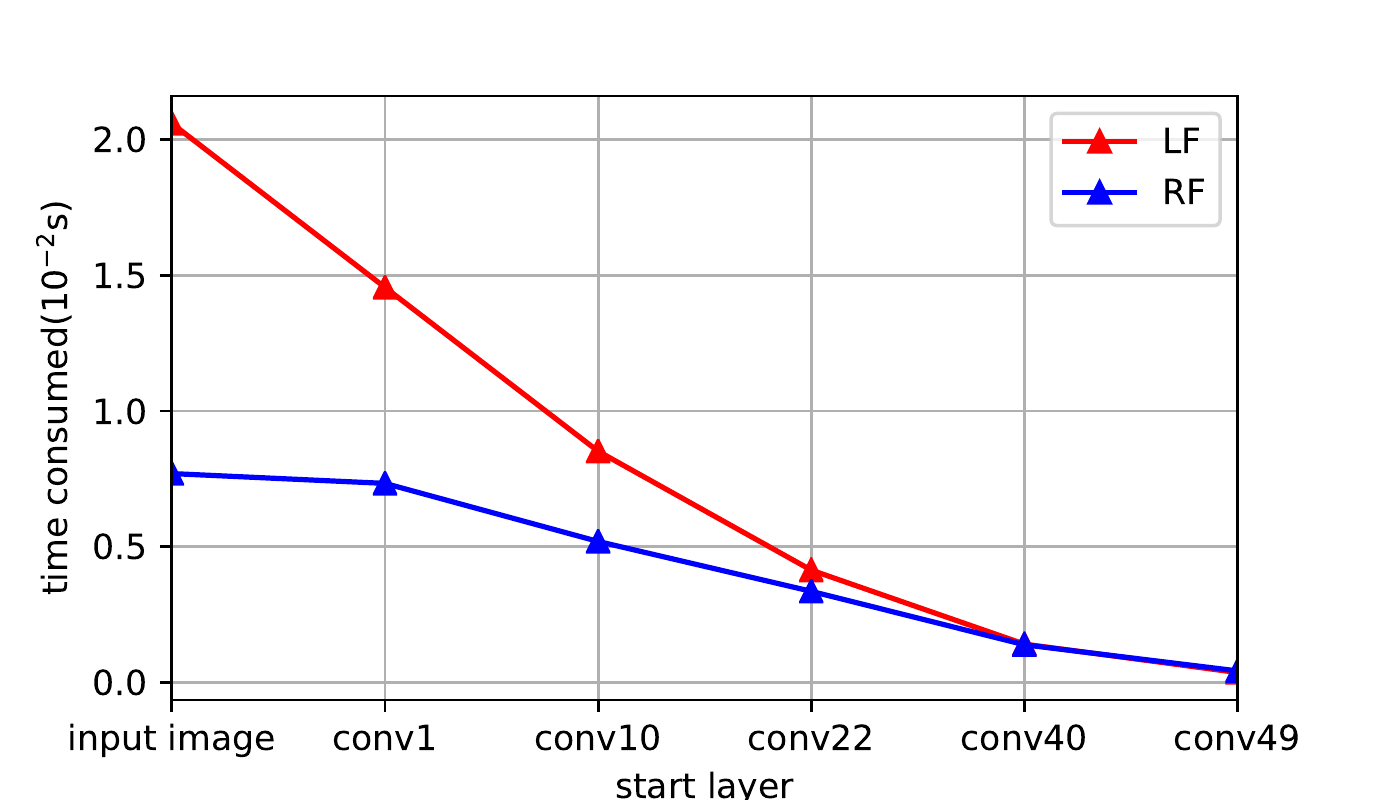}
\caption{Benchmark results for LF and RF. The x-axis shows the earliest layer used, after which all the layers are updated.
RF becomes efficient as more layers are used.
}
\label{fig:time}
\end{figure}

\section{What Layers are the Most Useful?}
\label{sec:layer-by-layer}
\vspace{-0.05in}
In this section, we analyze where are the most useful intermediate representations in a CNN under feedback-prop. In other words, what are the intermediate layers of a CNN that seem to allow maximal sharing of information among target predictions.
We first train three multi-label models based on Resnet-18, Resnet-50, and VGG-16 on the COCO multi-label task from Sec~\ref{sec:multi-label-experiment}. For each model we report in tables \ref{tab:layer-analysis-3}, \ref{tab:layer-analysis-1}, and \ref{tab:layer-analysis-2} the best validation accuracy that can be reached with the outputs of several individual layers as pivots using both LF and RF. We observe that in both VGG and Resnets, middle layers seem to be the most useful compared to layers closer to inputs or outputs. Specifically, we find that Conv13 in Resnet-18, Conv20 and Conv40 in Resnet-50, and Conv7 and Conv10 in VGG-16 achieve the best performance given the same amount of partial evidence (a fixed set of 500 \emph{known} labels and 500 \emph{unknown} labels). These results seem analogous to a recent study on neural networks where mutual information between intermediate representations with respect to both inputs and outputs is analyzed during training~\cite{shwartz2017opening}.
It would be interesting to devise an approach to automatically identify what layers are most effective to use as pivots under feedback-prop using an information theoretic approach.

\vspace{-0.06in}
\section{Computational Efficiency}
\label{sec:residual-efficient}
Here, we benchmark our two proposed feedback-prop methods. We use $\text{Resnet-50}$ multi-label model of Sec~\ref{sec:multi-label-experiment} and select a sequence of layers including \emph{input image}, \emph{conv1}, \emph{conv10}, \emph{conv22}, \emph{conv40}, and \emph{conv49}. We pick one layer as initial layer and update this layer with all subsequent layers. For example, if \emph{conv40} is the initial layer, we also update \emph{conv49}. We use a single 12GB NVIDIA Pascal Titan X GPU and record average inference times per image per iteration. Figure \ref{fig:time} shows that as more layers are used as pivots, RF shows the more gains over LF. RF is generally faster, with a slight increase in memory footprint.

\section{Conclusions}
In the context of deep CNNs, we found that by optimizing the intermediate representations for a given input sample during inference with respect to a subset of the target variables, predictions for all target variables improve their accuracy. We proposed two variants of a feedback propagation inference approach to leverage this dynamic property of CNNs and showed their effectiveness for making predictions under partial evidence for general CNN models trained in a multi-label or multi-task setting. As multi-task models trained to solve a wide array of tasks such as UberNet~\cite{Kokkinos17UberNet} emerge, we expect a technique such as feedback-prop will become increasingly useful. An interesting future direction would be devising an approach that leverages feedback-based updates during training. 

\noindent {\bf Acknowledgements}
This work was partially supported by a Google Faculty Research Award in Machine Perception.

{\small
\bibliographystyle{ieee}
\bibliography{egbib}

\begin{thebibliography}{10}\itemsep=-1pt

\bibitem{Agrawal2014AnalyzingTP}
P.~Agrawal, R.~B. Girshick, and J.~Malik.
\newblock Analyzing the performance of multilayer neural networks for object
  recognition.
\newblock In {\em ECCV}, 2014.

\bibitem{bar2004visual}
M.~Bar.
\newblock Visual objects in context.
\newblock {\em Nature Reviews Neuroscience}, 5(8):617--629, 2004.

\bibitem{netdissect2017}
D.~Bau, B.~Zhou, A.~Khosla, A.~Oliva, and A.~Torralba.
\newblock Network dissection: Quantifying interpretability of deep visual
  representations.
\newblock In {\em CVPR}, 2017.

\bibitem{biederman1982scene}
I.~Biederman, R.~J. Mezzanotte, and J.~C. Rabinowitz.
\newblock Scene perception: Detecting and judging objects undergoing relational
  violations.
\newblock {\em Cognitive psychology}, 14(2):143--177, 1982.

\bibitem{Binder2016}
A.~Binder, G.~Montavon, S.~Bach, K.-R. M{\"{u}}ller, and W.~Samek.
\newblock {Layer-wise Relevance Propagation for Neural Networks with Local
  Renormalization Layers}.
\newblock {\em arXiv:1604.00825}, 2016.

\bibitem{cao2015look}
C.~Cao, X.~Liu, Y.~Yang, Y.~Yu, J.~Wang, Z.~Wang, Y.~Huang, L.~Wang, C.~Huang,
  W.~Xu, et~al.
\newblock Look and think twice: Capturing top-down visual attention with
  feedback convolutional neural networks.
\newblock In {\em ICCV}, pages 2956--2964, 2015.

\bibitem{chun1998contextual}
M.~M. Chun and Y.~Jiang.
\newblock Contextual cueing: Implicit learning and memory of visual context
  guides spatial attention.
\newblock {\em Cognitive psychology}, 36(1):28--71, 1998.

\bibitem{meteor}
M.~Denkowski and A.~Lavie.
\newblock Meteor universal: Language specific translation evaluation for any
  target language.
\newblock In {\em In Proceedings of the Ninth Workshop on Statistical Machine
  Translation}, 2014.

\bibitem{santosh}
S.~K. Divvala, D.~Hoiem, J.~H. Hays, A.~A. Efros, and M.~Hebert.
\newblock An empirical study of context in object detection.
\newblock In {\em CVPR}, pages 1271--1278. IEEE, 2009.

\bibitem{escorcia2015relationship}
V.~Escorcia, J.~C. Niebles, and B.~Ghanem.
\newblock On the relationship between visual attributes and convolutional
  networks.
\newblock In {\em CVPR}, pages 1256--1264, 2015.

\bibitem{fang2015captions}
H.~Fang, S.~Gupta, F.~N. Iandola, R.~K. Srivastava, L.~Deng, P.~Doll{\'a}r,
  J.~Gao, X.~He, M.~Mitchell, J.~C. Platt, C.~L. Zitnick, and G.~Zweig.
\newblock From captions to visual concepts and back.
\newblock In {\em CVPR}, pages 1473--1482, 2015.

\bibitem{galleguillos2008object}
C.~Galleguillos, A.~Rabinovich, and S.~Belongie.
\newblock Object categorization using co-occurrence, location and appearance.
\newblock In {\em CVPR}, pages 1--8. IEEE, 2008.

\bibitem{gatys2016image}
L.~A. Gatys, A.~S. Ecker, and M.~Bethge.
\newblock Image style transfer using convolutional neural networks.
\newblock In {\em CVPR}, pages 2414--2423, 2016.

\bibitem{goodfellow2014generative}
I.~Goodfellow, J.~Pouget-Abadie, M.~Mirza, B.~Xu, D.~Warde-Farley, S.~Ozair,
  A.~Courville, and Y.~Bengio.
\newblock Generative adversarial nets.
\newblock In {\em NIPS}, pages 2672--2680, 2014.

\bibitem{goodfellow2014explaining}
I.~J. Goodfellow, J.~Shlens, and C.~Szegedy.
\newblock Explaining and harnessing adversarial examples.
\newblock {\em arXiv preprint arXiv:1412.6572}, 2014.

\bibitem{resnet}
K.~He, X.~Zhang, S.~Ren, and J.~Sun.
\newblock Deep residual learning for image recognition.
\newblock {\em CVPR}, 2016.

\bibitem{hu2016learning}
H.~Hu, G.-T. Zhou, Z.~Deng, Z.~Liao, and G.~Mori.
\newblock Learning structured inference neural networks with label relations.
\newblock In {\em CVPR}, pages 2960--2968, 2016.

\bibitem{johnson2015love}
J.~Johnson, L.~Ballan, and F.-F. Li.
\newblock Love thy neighbors: Image annotation by exploiting image metadata.
\newblock {\em ICCV}, 2015.

\bibitem{Karpathy_2015_CVPR}
A.~Karpathy and L.~Fei-Fei.
\newblock Deep visual-semantic alignments for generating image descriptions.
\newblock In {\em The IEEE Conference on Computer Vision and Pattern
  Recognition (CVPR)}, June 2015.

\bibitem{Kokkinos17UberNet}
I.~Kokkinos.
\newblock Ubernet: Training a 'universal' convolutional neural network for
  low-, mid-, and high-level vision using diverse datasets and limited memory.
\newblock {\em CVPR}, 2017.

\bibitem{alexnet}
A.~Krizhevsky, I.~Sutskever, and G.~E. Hinton.
\newblock Imagenet classification with deep convolutional neural networks.
\newblock In {\em Neural Information Processing Systems (NIPS)}, pages
  1097--1105, 2012.

\bibitem{mscoco}
T.-Y. Lin, M.~Maire, S.~Belongie, J.~Hays, P.~Perona, D.~Ramanan,
  P.~Doll{\'a}r, and C.~L. Zitnick.
\newblock Microsoft coco: Common objects in context.
\newblock In {\em ECCV}, pages 740--755. Springer, 2014.

\bibitem{roozbeh}
R.~Mottaghi, X.~Chen, X.~Liu, N.-G. Cho, S.-W. Lee, S.~Fidler, R.~Urtasun, and
  A.~Yuille.
\newblock The role of context for object detection and semantic segmentation in
  the wild.
\newblock In {\em CVPR}, June 2014.

\bibitem{murphy1999loopy}
K.~P. Murphy, Y.~Weiss, and M.~I. Jordan.
\newblock Loopy belief propagation for approximate inference: An empirical
  study.
\newblock In {\em Proceedings of the Fifteenth conference on Uncertainty in
  artificial intelligence}, pages 467--475. Morgan Kaufmann Publishers Inc.,
  1999.

\bibitem{navon1977forest}
D.~Navon.
\newblock Forest before trees: The precedence of global features in visual
  perception.
\newblock {\em Cognitive psychology}, 9(3):353--383, 1977.

\bibitem{palmer1975effects}
S.~E. Palmer.
\newblock The effects of contextual scenes on the identification of objects.
\newblock {\em Memory \& Cognition}, 3:519--526, 1975.

\bibitem{bleu}
K.~Papineni, S.~Roukos, T.~Ward, and W.-J. Zhu.
\newblock Bleu: A method for automatic evaluation of machine translation.
\newblock In {\em Proceedings of the 40th Annual Meeting on Association for
  Computational Linguistics}, ACL '02, pages 311--318, Stroudsburg, PA, USA,
  2002. Association for Computational Linguistics.

\bibitem{rabinovich2007objects}
A.~Rabinovich, A.~Vedaldi, C.~Galleguillos, E.~Wiewiora, and S.~Belongie.
\newblock Objects in context.
\newblock In {\em ICCV}, pages 1--8. IEEE, 2007.

\bibitem{Ruan2017Ensemble}
W.~Ruan and E.~L. Miller.
\newblock Ensemble multi-task gaussian process regression with multiple latent
  processes.
\newblock {\em arXiv}, 2017.

\bibitem{rumelhart}
D.~E. Rumelhart, G.~E. Hinton, and R.~J. Williams.
\newblock Learning representations by back-propagating errors.
\newblock {\em Cognitive modeling}, 5(3):1, 1988.

\bibitem{Salakhutdinov2009DeepBM}
R.~Salakhutdinov and G.~E. Hinton.
\newblock Deep boltzmann machines.
\newblock In {\em AISTATS}, 2009.

\bibitem{selvaraju2016grad}
R.~R. Selvaraju, A.~Das, R.~Vedantam, M.~Cogswell, D.~Parikh, and D.~Batra.
\newblock Grad-cam: Why did you say that? visual explanations from deep
  networks via gradient-based localization.
\newblock {\em arXiv preprint arXiv:1610.02391}, 2016.

\bibitem{shwartz2017opening}
R.~Shwartz-Ziv and N.~Tishby.
\newblock Opening the black box of deep neural networks via information.
\newblock {\em arXiv preprint arXiv:1703.00810}, 2017.

\bibitem{simonyan2013deep}
K.~Simonyan, A.~Vedaldi, and A.~Zisserman.
\newblock Deep inside convolutional networks: Visualising image classification
  models and saliency maps.
\newblock {\em arXiv preprint arXiv:1312.6034}, 2013.

\bibitem{vgg}
K.~Simonyan and A.~Zisserman.
\newblock Very deep convolutional networks for large-scale image recognition.
\newblock {\em ICLR}, 2015.

\bibitem{cider}
R.~Vedantam, C.~L. Zitnick, and D.~Parikh.
\newblock Cider: Consensus-based image description evaluation.
\newblock {\em CoRR}, abs/1411.5726, 2014.

\bibitem{showandtell}
O.~Vinyals, A.~Toshev, S.~Bengio, and D.~Erhan.
\newblock Show and tell: A neural image caption generator.
\newblock In {\em CVPR}, 2015.

\bibitem{vittayakorn2016automatic}
S.~Vittayakorn, T.~Umeda, K.~Murasaki, K.~Sudo, T.~Okatani, and K.~Yamaguchi.
\newblock Automatic attribute discovery with neural activations.
\newblock In {\em ECCV}, pages 252--268. Springer, 2016.

\bibitem{Wang2017Attribute}
J.~Wang, X.~Zhu, S.~Gong, and W.~Li.
\newblock Attribute recognition by joint recurrent learning of context and
  correlation.
\newblock {\em arXiv}, 2017.

\bibitem{wu2013tag}
L.~Wu, R.~Jin, and A.~K. Jain.
\newblock Tag completion for image retrieval.
\newblock {\em IEEE Transactions on Pattern Analysis and Machine Intelligence},
  35(3):716--727, 2013.

\bibitem{xiao2010sun}
J.~Xiao, J.~Hays, K.~A. Ehinger, A.~Oliva, and A.~Torralba.
\newblock Sun database: Large-scale scene recognition from abbey to zoo.
\newblock In {\em CVPR}, pages 3485--3492. IEEE, 2010.

\bibitem{xie2016interactive}
L.~Xie, L.~Zheng, J.~Wang, A.~L. Yuille, and Q.~Tian.
\newblock Interactive: Inter-layer activeness propagation.
\newblock In {\em CVPR}, pages 270--279, 2016.

\bibitem{zeiler2014visualizing}
M.~D. Zeiler and R.~Fergus.
\newblock Visualizing and understanding convolutional networks.
\newblock In {\em ECCV}, pages 818--833. Springer, 2014.

\bibitem{zhang2016top}
J.~Zhang, Z.~Lin, S.~X. Brandt, Jonathan, and S.~Sclaroff.
\newblock Top-down neural attention by excitation backprop.
\newblock In {\em ECCV}, 2016.

\bibitem{zhou2017places}
B.~Zhou, A.~Lapedriza, A.~Khosla, A.~Oliva, and A.~Torralba.
\newblock Places: A 10 million image database for scene recognition.
\newblock {\em IEEE Transactions on Pattern Analysis and Machine Intelligence},
  2017.

\end{thebibliography}
}

\end{document}